\documentclass{article} % For LaTeX2e
\usepackage[preprint]{colm2025_conference} %final

\usepackage{microtype}
\usepackage{hyperref}
\usepackage{url}
\usepackage{booktabs}
\usepackage{graphicx} 
\usepackage{lineno}
\usepackage{adjustbox}
\usepackage{enumitem}

\definecolor{darkblue}{rgb}{0, 0, 0.5}
\hypersetup{colorlinks=true, citecolor=darkblue, linkcolor=darkblue, urlcolor=darkblue}

\title{Non-Determinism of ``Deterministic'' LLM Settings} % Impairs Evaluation}

\author{Berk Atil$^1$\thanks{Berk Atil completed this work during his internship at Comcast AI Technologies} , Sarp Aykent$^2$, Alexa Chittams$^2$, Lisheng Fu$^2$, Rebecca J. Passonneau$^1$, \\
\textbf{Evan Radcliffe$^2$, Guru Rajan Rajagopal$^2$, Adam Sloan$^2$, Tomasz Tudrej$^2$, Ferhan Ture$^2$,} \\
\textbf{Zhe Wu$^2$, Lixinyu Xu$^2$, Breck Baldwin$^2$}  \\ \\
\textsuperscript{\rm 1}Penn State University, \textsuperscript{\rm 2}Comcast AI Technologies\\ \\
\textbf{Correspondence:} \texttt{\{bka5352,rjp49\}@psu.edu}; \texttt{breckbaldwin@gmail.com} 
}

\begin{document}

\ifcolmsubmission
\linenumbers
\fi

\maketitle

\begin{abstract}
LLM (large language model) practitioners commonly notice that outputs can vary for the same inputs under settings expected to be deterministic. %, but we have been unable to find work that evaluates LLM stability as the main objective. 
Yet the questions of how pervasive this is, and with what impact on results, have not to our knowledge been %directly addressed in a 
systematically investigated.
We investigate non-determinism in five LLMs configured to be deterministic when applied to eight common tasks in across 10 runs, in both zero-shot and few-shot settings. We see accuracy variations up to 15\% across naturally occurring runs with a gap of best possible performance to worst possible performance up to 70\%. In fact, none of the LLMs consistently delivers repeatable accuracy across all tasks, much less identical output strings. 
%Interestingly, these variations in accuracy are not normally distributed. %We also compare configurations with zero-shot or few-shot prompting and fine-tuning. 
Sharing preliminary results with insiders has revealed that non-determinism perhaps essential to the efficient use of compute resources via co-mingled data in input buffers so this issue is not going away anytime soon. 
To better quantify our observations, we introduce metrics focused on %outlier results:
quantifying determinism, TARr@N for the total agreement rate at N runs over raw output, and TARa@N for total agreement rate of parsed-out answers. Our code and data are publicly available at https://github.com/breckbaldwin/llm-stability. %We suggest that measures of uncertainty be integrated into leader boards and research results going forward.
\end{abstract}

\section{Introduction}

%Large Language Models (LLM) have increasingly penetrated many application areas, %with their powerful capabilities including question answering \citep{robinson2023leveraging}, reasoning \citep{qiao-etal-2023-reasoning}, code generation \citep{jiang2024survey} in NLP, and many other high stakes applications in health care, crime control, and epidemiology.  

%Alongside 
Large Language Models (LLM) perform well on many types of NLP or NLP-related tasks, including question answering \citep{robinson2023leveraging}, diverse types of reasoning \citep{qiao-etal-2023-reasoning}, code generation \citep{jiang2024survey}, and the like. Given their general applicability and their widespread adoption for diverse, high-stakes societal functions, such as information gathering in medicine \citep{shool2025systematic} or law \citep{niklaus-etal-2024-multilegalpile}, financial planning \citep{de2024optimized}, or manufacturing optimization \citep{du2025llm} to name a few, there has been increasing attention to reliability (e.g., for Out-of-Distribution behavior \citep{liu-etal-2024-good,du2022towards}), alongside other aspects of LLM trustworthiness \citep{shridhar-etal-2024-art,chen-mueller-2024-quantifying}. A laudable dimension of the LLM field as a whole is the broad use of benchmark datasets: you know what you are buying against an objective metric. However, what is the value of such benchmarks if LLM results are not stable across multiple runs? One aspect of reliability that has not to our knowledge been addressed
sufficiently in the literature is lack of user control over variance in LLM output across LLMs
and platforms under settings that are assumed to be deterministic. We demonstrate an alarming degree of variation across equivalent input runs with a varied collection of high performing LLMs under presumed deterministic settings. Our findings suggest %luck may have far to great a role 
there is far too much uncertainty in a realm where robust engineering %information
is the expectation. %called for. 

%One aspect of reliability that has not to our knowledge been addressed sufficiently in the literature is lack of user control over variance in LLM output across LLMs and platforms, even when the user expects deterministic performance. In our investigation of the primary method to control determinism across multiple LLMs on diverse tasks, we find sizable variance.

Besides training procedure and data, hyper-parameters such as temperature, top-k, top-p, and repetition penalty can affect model performance significantly \citep{wang2023cost}. Top-k and top-p shrink sampling space based on the probabilities, whereas temperature modifies the probability distribution over the vocabulary \citep{wang2020contextual,wang2023cost}. When the temperature is high, the distribution flattens; however, when it is closer to 0, the model is supposed to be more deterministic. %The naive expectation would be that the model is entirely deterministic with temperature=0.
Many users of LLM platforms expect model performance to be deterministic when temperature=0. While many other users may have observed a degree of non-determinism in this setting, there is little if any quantification of this variance. We refer here to this behavior of output variance despite zero temperature as instability.\footnote{In this paper, we use stability/instability and determinism/non-determinism interchangeably.}

Despite known uncertainty across different training runs of neural networks, it has become standard to measure LLM performance on benchmark datasets %concludes 
by reporting a single result %output 
\citep{hendrycks2021measuring,suzgun-etal-2023-challenging,wang2024mmlu,gema2024we,rein2023gpqa}, possibly due to cost and computational time restrictions. However, if there is a significant variance in the output across identical training runs, this reduces the validity of the benchmarks %since 
given that measures of uncertainty (e.g. confidence intervals) are often not included. %reported result may just be due to random chance. 
Here we look at an additional source of uncertainty, performance variance from the same trained model on the same inputs when temperature=0.

\begin{figure}[t]
    \centering
    \includegraphics[width=\textwidth]{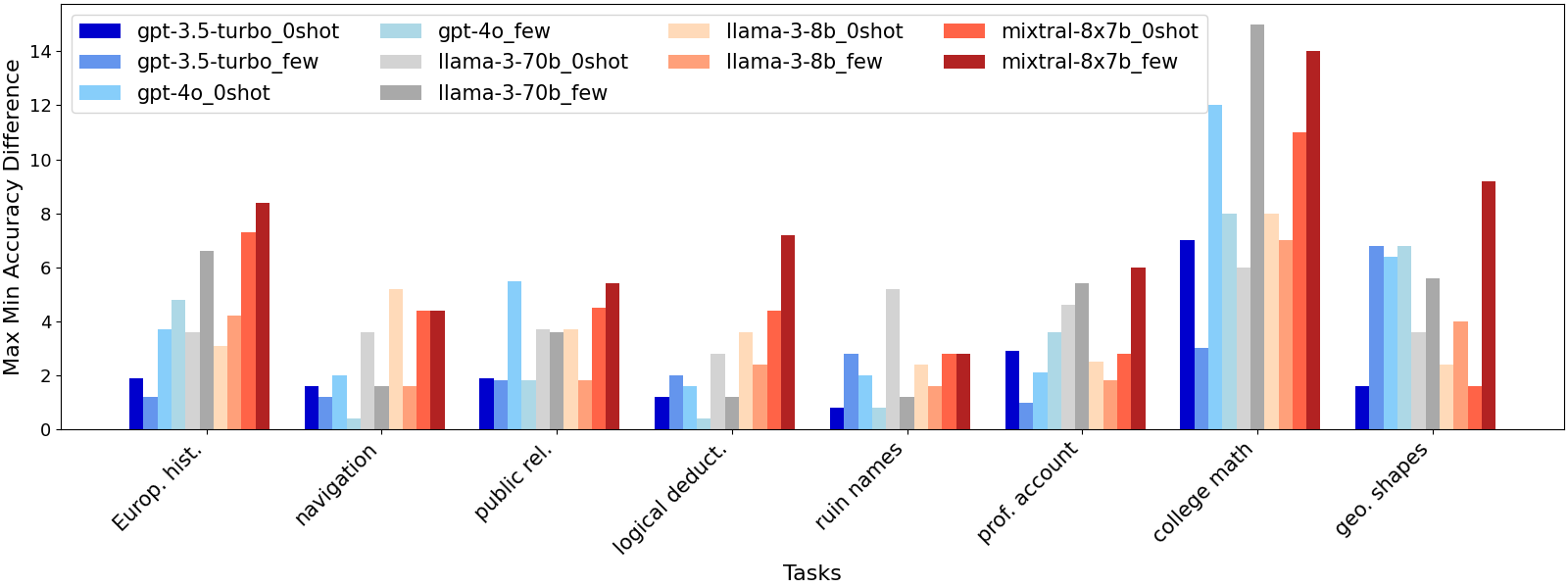}
    \caption{Percentage difference between maximum and minimum accuracy in 10 runs per model, for 5 models on 8 tasks with zero-shot and few-shot settings. % in terms of \%.
    }
    \label{fig:acc_diff}
\end{figure}

In this work, we analyze the instability of various families of models on tasks from two common benchmarks: Massive Multititask Language Understanding (MMLU) \citep{hendrycks2021measuring}) and BIG-Bench Hard (BBH) \citep{suzgun-etal-2023-challenging}). Figure~\ref{fig:acc_diff} shows that models are not deterministic, and that the degree of instability changes from model to model, setting to setting, and task to task. Therefore, performance instability can doubtless impact the ranked performance of models. % across identical input runs. 
Our specific contributions include:
\begin{itemize}[itemsep=0pt]
    \item Two metrics, TARr@N (total agreement rate for raw data across N runs) and TARa@N (total agreement rate for parsed answer across N runs) for LLM instability to capture the variability in answer accuracy and in the output word spans.
    \item Quantification of LLM instability over 8 tasks randomly selected from two common benchmarks: BBH and MMLU.
    \item Comparison across settings, including zero-shot and few-shot (3 for BBH, 5 for MMLU as in the standard settings).
    \item Correlation analysis of instability with accuracy, input length, and output length.
    \item Data from runs and source code.\footnote{Download URL will be made available if paper is accepted} %are available at https://github.com/Comcast/llm-stability.
\end{itemize}

%We show that a significant LLM output variability might occur even with the most ``deterministic'' settings (e.g. hyperparameters such as temperature). This exists for 5 models from different families on %two datasets totaling 8 tasks. 
%eight tasks selected from MMLU and BBH.
%As shown in Figure~\ref{fig:acc_diff}, we find accuracy differences for the same model and task of up to 15\% across ten runs. 
Our systematic investigation is structured into the following sections:
%The paper sections in order are 
Introduction, Related Work, Controlling LLM Determinism, Datasets, Methods, Experiments, Results, Discussion, and Conclusion.

%% Please note that we have introduced automatic line number generation
%% into the style file for \LaTeXe. This is to help reviewers
%% refer to specific lines of the paper when they make their comments. Please do
%% NOT refer to these line numbers in your paper as they will be removed from the
%% style file for the final version of accepted papers.

\section{Related Work}

There have been investigations on the robustness of machine learning (ML) models with trivial changes to the input \citep{10.1145/3338501.3357372,freiesleben2023beyond,10.1145/3351095.3372836,rauber2017foolbox}. However, we are interested in the changes in the response when everything in the input is the same.
%More recently, similar analyses also show the unreliability of the benchmarks and robustness of LLMs \citep{xu2024llms,gupta2024changing,sinha-etal-2023-language,sarker2024syntacticrobustnessllmbasedcode,raina2024llm}. 
%For multiple choice questions, the order of the options changes the performance of the models significantly \citep{gupta2024changing}. \citet{xu2024llms} replace the correct option with ``none of the above'' and observe a dramatic performance change across different models. However, our focus is on the exact same input and setup, and noting the variations across N runs.
\citet{biderman2024lessons} introduce a standard evaluation toolkit for LLMs and suggest best practices for reproducibility; instability, however, is not addressed. Works that mention instability include \citep{song2024good}, which analyzes the effect of temperature, sampling strategy, repetition penalty, and alignment algorithms on the performance. %They have similar findings 
While this work finds
that LLMs have some variance in the output that should be taken into account in the evaluation benchmarks, they use a temperature of 1, %when they report the variance of the outputs, 
thereby introducing the variability that our study seeks to minimize. %eliminate. 
\citet{ouyang2023llm} do a similar instability analysis of a single model, ChatGPT, with varying temperatures on a single type of task, code generation. Lastly, \citet{Holtzman2020The} mention the freedom in text generation which might lead to different outputs for the same inputs, but they do not talk about the parameters that affect this behaviour. To the best of our knowledge, no work systematically investigates LLM instability given the same inputs and configurations (zero-shot and few-shot) with maximally deterministic hyper-parameters. % multiple times to assess the instability of LLM output. 

\section{Controlling LLM Determinism}

Determinism is an important yet often overlooked challenge, with variations in practice across platforms. %vendors and implementations. 
%There are some strong speculations, most informatively delivered by ad-hoc means such as forum posts \citep{openai_cheat_sheet}. 
Temperature is the key parameter, with 0 being the most deterministic value and 1 the least. %It controls the ``creativity'' via the softmax function which changes the determinism of the output significantly. 
Equation \ref{eqn:temp} shows the probability of word $i$ where $T$ is temperature and $y_i$ is the LLM logit.

\vspace*{-.2in}
\begin{equation}  
\frac{e^{\frac{y_i}{T}}}{\sum_{j=1}^Ne^{\frac{y_j}{T}}} 
\label{eqn:temp}
\end{equation} 

When $T$ is set to 0, the LLM should theoretically produce the same output given the same prompt. %, on the other hand, it makes the model less creative. 
$T$ can be raised to diversify outputs.
%We also experimented with top-p, which keeps the smallest set of tokens with probabilities that add up to set value, but found no difference for our stability purposes, so we do not report it.
We confirmed that other potentially relevant hyper-parameters, such as top-p (the smallest set of tokens with probabilities summing to $p$) have little effect on determinism. %do not affect determinism much.
%There may be other ways to achieve determinism for any given model, such as greedy decoding, but our goal is to model a fairly standard evaluation that can be used in leaderboards or comparing the performance of different approaches. 

% \begin{figure}[t]
%     \centering
%     \includegraphics[width=0.6\textwidth]{20_trial_1.png}
%     \caption{Accuracy over 20 identical runs on college math, temperature=0, top-p=1. Median in blue, mean in black with dashed 5\% and 95\% quantiles.}
%     \label{fig:twenty_zero_one}
% \end{figure}

\section{Datasets}

To ensure that our investigation of instability includes diverse Natural Language Processing (NLP) tasks, we turned to two widely used multiple-choice benchmarks: Beyond the Imitation Game Benchmark Hard (BBH) \citep{suzgun-etal-2023-challenging}, with 27 diverse tasks from mathematics, commonsense reasoning and other domains; Measuring Massive Multitask Language Understanding (MMLU) \citep{hendrycks2021measuring}, with 57 tasks across disciplines including the humanities, social sciences, and STEM areas. To balance diversity against computational resources, we randomly selected four subtasks from each benchmark. Table \ref{tab:dataset_stats} lists the tasks, number of examples, and number of multiple-choice options.

%Beyond Imitation Game Benchmark Hard (BBH) \citep{suzgun-etal-2023-challenging} is a benchmark consisting of 27 challenging tasks about traditional Natural Language Processing (NLP), mathematics, commonsense reasoning, etc. We randomly selected ``navigation'', a task to determine whether or not an agent returns back to the starting point given navigational steps; ``ruin names'', a task to pick a humorous simple edit of a band or movie name; ``geometric shapes'', a task to determine the geometric shape given in SVG path format; and ``logical deduction with three objects'', a task to deduce the order of a sequence of three objects from a set of conditions. 

%Measuring Massive Multitask Language Understanding (MMLU) \citep{hendrycks2021measuring} is another benchmark that contains 57 tasks in humanities, social sciences, STEM, and other important fields to learn. We randomly picked ``high school European history'' from humanities, with questions about the history of Europe; ``college mathematics'' from STEM with questions about calculus, algebra etc.; ``public relations'' from social sciences with questions about media theory, crisis management etc.; and ``professional accounting'' with accounting questions. All of the tasks are multiple choice questions with varying number of options, see brief statistics in Table \ref{tab:dataset_stats}.

\begin{table}[t]
\begin{center}
\begin{tabular}{l|l|r|r}\hline 
%\textbf{Task} & \textbf{\# of Examples} & \textbf{\# of Options} \\ \hline
\multicolumn{1}{c}{Task} & 
    \multicolumn{1}{|c}{Descrip.} & \multicolumn{1}{|c}{Size} &
        \multicolumn{1}{|c}{Options} \\\hline
BBH: navigation & does path end at start & 250 &  2\\
BBH: ruin names & humorous edit of a band or movie title & 250 & 4 \\
BBH: geometric shapes & shape given SVG format & 250 & 10 \\
BBH: logical deduct. 3 objects & order of 3 objects given constraints & 250 & 3 \\
MMLU: h. s. Europ. hist. & \textit{identical} & 165 &  4\\
MMLU: college math & \textit{identical} & 100 & 4 \\
MMLU: prof. accounting &\textit{identical} & 282  & 4  \\
MMLU: public rel. & media theory, crisis mgmt., etc.& 110 & 4\\
%beat 1 & 100 & 2\\
%beat 2 & 100 & 2\\
%beat 3 & 100 & 2\\
\hline
\end{tabular}
\end{center}
%\caption{\label{tab:dataset_stats} Statistics about the tasks.}
\caption{Eight tasks from BBH and MMLU with brief descriptions, number of examples, and number of answer options.}
\label{tab:dataset_stats} 
\end{table}

\section{Methods}

For our investigation of instability, or non-determinism of zero-temperature settings, we 
perform experiments on models without fine-tuning in
both zero-shot and few-shot prompting (without Chain-of-Thought (CoT) \citep{wei2022chain}). Regarding the number of examples for few-shot, we use the standard settings of 3-shot for BBH tasks, and 5-shot for MMLU tasks. We set the temperature at 0, top-p at 1, and fix the seed. We use the same compute infrastructure, inputs, and configurations.

\begin{figure}[t]
    \centering
    \includegraphics[width=0.6\textwidth]{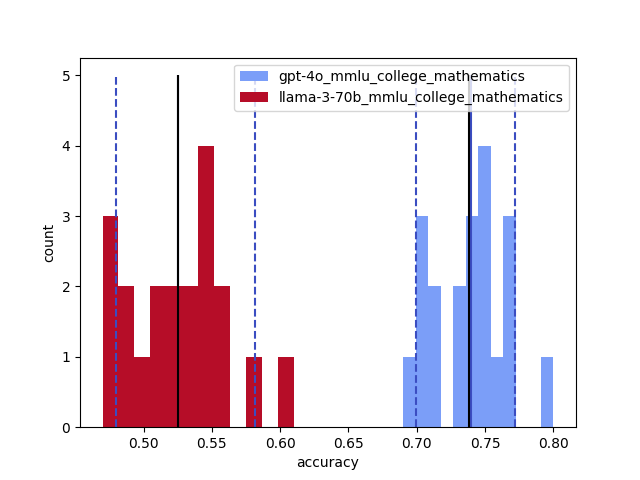}
    \caption{Accuracy over 20 identical runs on college math, temperature=0, top-p=1. Median in blue, mean in black with dashed 5\% and 95\% quantiles.}
    \label{fig:twenty_zero_one}
\end{figure}

\subsection{Models}

We chose five top performing models from different families and with varying sizes: GPT-3.5 Turbo \citep{brown2020language}, GPT-4o \citep{openai2024gpt4technicalreport}, Llama-3-70B-Instruct \citep{llama3}, Llama-3-8B-Instruct \citep{llama3}, and Mixtral-8x7B-Instruct \citep{jiang2024mixtralexperts}.

% We experiment with GPT-3.5 Turbo \citep{brown2020language}, GPT-4o \citep{openai2024gpt4technicalreport}, Llama-3-70B-Instruct \citep{llama3}, Llama-3-8B-Instruct \citep{llama3}, and Mixtral-8x7B-Instruct \citep{jiang2024mixtralexperts} because they were some of the top performing models from different families and with varying sizes.

\subsection{Metrics}

To quantify instability, we collect  accuracy across a set of runs in the same condition (model $\times$ dataset). % to characterize the impact of LLM instability on apparent performance. 
We also define five metrics specifically to quantify instability for each condition that capture the extremes of positive and negative accuracy. In addition, we report median accuracy to see the deviation from median as well. Note that we do not report standard deviations of means because 
%we are interested in assessing best and worst cases rather than an expectation; further,
the distributions across runs for a given condition are not normally distributed (see below).
%The focus of this work is evaluating the determinism of LLMs across tasks, not the performance, but we report accuracy to validate that the LLMs are performing as expected. Mainly, we measure accuracy variation across runs which characterizes the larger impact of ignoring LLM instability when evaluating them. We also define some instability metrics that concern the reproducibility of answers and raw output.
\begin{itemize}
    %\item Minimum-maximum spread, which is the difference between the minimum and maximum accuracy over the runs.
    \item Total agreement rate@N (TAR@N): the percentage of test set questions across N runs where generated answers are all identical, \textit{regardless of whether the answer was correct}. %Its value might vary depending on the number of runs so we have @N notation. 
    We have two variants of TAR@N:
    \begin{itemize}
        \item  TARr@N (TAR@N for the raw model response) The LLM responses are string equivalent.% The parsed answer extracted from the response may or may not be correct.
        \item TARa@N (TAR@N for the answer) The parsed answers are the same, e.g., ``The answer is a)'' is the same as ``a) is the answer''. %The answer may or may not be correct.
    \end{itemize} 
    \item The best possible accuracy over $N$ runs (BestAcc), which is the maximum possible accuracy %that you can get over N runs. 
    that could be extracted from $N$ runs.
    For each question, if there is a run in which %any one of 
    the answer is correct, that question is marked as correctly answered.
    \item The worst possible accuracy over $N$ runs (WorstAcc), which is the minimum possible accuracy %that you can get over N runs. 
    that could be extracted from $N$ runs.
    For each question, if there is a run in which the %any one of the
    answer is incorrect, that question is marked as incorrectly answered.
     \item Median accuracy over N runs.
     \item Maximum-minimum accuracy difference over N runs. Note that this is different from BestAcc and WorstAcc because we look at the actual accuracy scores for each run instead of focusing on worst case and best case scenarios.
    %\item Minimum, median, and maximum accuracy values over the runs.
\end{itemize}
% \begin{figure*}[t]
%     \centering    \includegraphics[width=\textwidth]{TAR_r_plot.png}
%     \caption{TARr@5 for each model, in terms of \%. All models are prompted in few-shot setting. When the colors change from dark red to dark blue, TARr@5 gets better.
%     }
%     \label{fig:TARr_medians}
% \end{figure*}

%Our metrics allow for the possibility of a 100\% mean accuracy which would have 100\% TARa@N but 0\% TARr@N score. 
TARr@N score is the strictest metric of instability, since any character variation will result in a disagreement. %The reason for not reporting standard deviation and means is that the variations are not normal. 
Thus it is possible for a set of runs to have 100\% accuracy and 100\% TARa@N combined with 0\% TARr@N.

To examine the distributional qualities of accuracy scores, we did 20 runs of GPT-4o and Llama-3-70b on college math in few-shot, which are  
two of the more unstable examples. %conditions. %configurations. , to do the normality tests. 
The results shown in Figure \ref{fig:twenty_zero_one} clearly show non-normal distributions, with mean and median values far from the mode. A Kolmogorov-Smirnov normality test \citep{massey1951kolmogorov} rejected the normal hypothesis with a p-value $< 10^{-9}$.

%\section{Experiments and Results}

\section{Experiments}

Our eight datasets, five base models and two settings (zero/few-shot) yield eighty conditions. For each condition, we perform ten runs with temperature at zero to quantify the amount of instability within a run. 

% \begin{figure}[t]
%     \centering
%     \includegraphics[width=0.7\textwidth]{corr_heat_map0shot.png}
%     \caption{Spearman correlation matrix between metrics in zero-shot setting for the models that show significant instability (all models except the fine-tuned model)}
%     \label{fig:corr_analysis_0shot}
% \end{figure}

\begin{table*}[t]
 \begin{center}
   \begin{adjustbox}{width=\textwidth}

 \begin{tabular}{l|c|c|c|c|c}
\hline \textbf{Task} & \textbf{gpt3.5} & \textbf{gpt4o}  & \textbf{llama8b} & \textbf{llama70b} & \textbf{mixtral8-7b}  \\ \hline 
\multicolumn{6}{c}{\textbf{Accuracy Results}} \\\hline
navigation & 96.8, 95.6, 93.2 & 98.8, 98.8, 98.4 & 82.0, 80.2, 78.0 & 95.2, 94.6, 93.6 & 84.4, 79.0, 71.6  \\
geo. shapes & 72.4, 59.6, 46.8 & 82.4, 68.4, 53.6 & 49.2, 40.6, 32.8 & 67.2, 57.0, 47.2 & 54.4, 27.8, 08.8  \\
logical deduct. & 88.8, 81.6, 75.2 & 100., 100., 99.6 & 95.6, 90.2, 81.2 & 98.0, 96.4, 95.2 & 87.6, 75.0, 64.0  \\
public rel. & 75.5, 69.1, 65.5 & 80.0, 76.4, 73.6 & 63.6, 61.8, 61.8 & 67.3, 60.5, 53.6 & 58.2, 48.2, 36.4  \\
Europ. hist. & 83.6, 81.2, 78.2 & 89.1, 81.5, 72.1 & 74.5, 67.0, 59.4 & 61.8, 50.3, 41.2 & 65.5, 51.5, 35.8  \\
ruin names & 72.0, 58.0, 44.8 & 93.2, 90.8, 88.4 & 68.4, 66.8, 64.4 & 89.2, 87.2, 84.4 & 78.8, 67.6, 55.6  \\
prof. account & 52.5, 50.9, 48.9 & 89.0, 74.5, 57.8 & 48.2, 45.4, 44.0 & 78.0, 67.2, 55.3 & 67.0, 39.0, 13.1  \\
college math & 39.0, 38.0, 34.0 & 88.0, 69.0, 44.0 & 50.0, 22.5, 04.0 & 85.0, 54.5, 22.0 & 75.0, 31.5, 03.0  \\
\hline  
\multicolumn{6}{c}{\textbf{TAR Results}} \\\hline
navigation & 96.4, 46.0 & 99.6, 46.0 & 96.0, 86.0 & 98.4, 64.0 & 84.8, 50.0  \\
geo. shapes & 62.8, 25.2 & 63.2, 00.0 & 58.8, 27.6 & 66.4, 18.0 & 12.0, 02.4  \\
logical deduct. & 84.4, 34.8 & 99.6, 36.8 & 85.2, 50.0 & 97.2, 49.6 & 74.8, 16.4  \\
public rel. & 87.3, 82.7 & 92.7, 37.3 & 96.4, 73.6 & 81.8, 17.3 & 62.7, 10.9  \\
Europ. hist. & 94.5, 70.9 & 81.2, 09.1 & 82.4, 07.3 & 73.3, 22.4 & 55.2, 23.6  \\
ruin names & 66.0, 05.6 & 95.2, 00.0 & 88.4, 47.6 & 94.4, 10.8 & 70.4, 24.8  \\
prof. account & 91.1, 76.6 & 66.7, 04.6 & 89.0, 52.1 & 69.5, 00.0 & 23.4, 00.7  \\
college math & 89.0, 76.0 & 50.0, 00.0 & 22.0, 00.0 & 25.0, 00.0 & 07.0, 00.0  \\
\hline
\end{tabular}
\end{adjustbox}
\end{center}
\caption{\label{tab:general_results} BestAcc, Median Accuracy, WorstAcc on top; TARa@10, TARr@10 on bottom, for the few-shot conditions (3 for BBH, 5 for MMLU, see section 5). Results are in terms of percentages.} %These models are prompted in few-shot setting.}
\end{table*}

% \begin{table}[t]
% \begin{center}
% \begin{adjustbox}{width=0.5\textwidth}
% \begin{tabular}{|l|c|c|}
% \hline \textbf{Model} & \textbf{TARa@5 Median} & \textbf{TARr@5 Median}  \\ \hline
% fine-tuned-gpt3.5-few & 100\% & 100\% \\
% gpt3.5-few & 99\% & 92\% \\
% gpt3.5-0shot & 99\% & 99\% \\
% mixtral8-7b-few & 89\% & 61\% \\
% mixtral8-7b-0shot & 99\% & 88\% \\
% llama8b-few & 92\% & 37\% \\
% llama8b-0shot & 100\% & 98\% \\
% llama70b-few & 93\% & 20\% \\
% llama70b-0shot & 69\% & 2\% \\
% gpt4o-few & 94\% & 3\% \\
% gpt4o-0shot & 93\% & 11\% \\
% \hline
% \end{tabular}
% \end{adjustbox}
% \end{center}
% \caption{\label{tab:model_rankings} Median of TARa@5 and TARr@5 over datasets. The models with the ``few'' in their name are prompted in few-shot setting.}
% \end{table}

\section{Results}

\subsection{Instability Results}

Figure \ref{fig:acc_diff} %differentiates experiment configurations such as fine-tuning, N-shot across tasks or LLMs for the minimum and maximum accuracy values. 
summarizes the extremes observed across our eight datasets for the five models in zero-shot and few-shot settings. The y-axis is the percentage difference between the minimum and maximum accuracies in ten runs for each condition.
Notably, there are 5-15\% differences on some tasks. %Many models have between 5-10\% spreads on some configuration/task.

%Table \ref{tab:general_results} summarizes the 5 run experiments across minimum, median, and maximum values followed by total agreement rate TARa@5 and TARr@5. Perfectly stable system results in the same score for minimum, median, and maximum accuracy, and 100\% for both TARa@5 and TARr@5. No model/task achieves this performance except the fine-tuned model on logical deduction and European history tasks.

The top of Table \ref{tab:general_results} reports BestAcc, median accuracy and WorstAcc in the few-shot conditions for our five models (zero-shot results show a similar degree of non-determinism, with varying consistency across conditions, see Table \ref{tab:general_results_0shot} in Appendix \ref{appendix:zero_shot}). The lower half of the table reports TARa@10 and TARr@10. %, \textcolor{red}{and the fine-tuned GPT-3.5}. 
%When the values for BestAcc,  WorstAcc and TARa@10 are relatively similar, models are relatively more stable.  Note that this  can occur for conditions that have relatively higher accuracy, shown by high values and small delta for BestAcc, WorstAcc (e.g., many models on Navigation), or relatively lower accuracy  (e.g., GPT-3.5 on European History). 
When there is a gap between BestAcc and WorstAcc $>10$ there is often very low TARr@10 (e.g., gpt3.5 on geometric shapes, logical deduction, ruin names; gpt4o on public relations, European history professional accounting, college math; and so on). 
Notably, TARr@10 is typically fairly low, and there is a lot of variation across models and datasets. Unsuprisingly, TARa@10 can be much higher than TARr@10 because of the flexibility of answer parsing that is not available for the raw outputs. Overall, models show a high degree of instability at both the raw output level but at the accuracy level as well. 

%Table \ref{tab:model_rankings} shows the median performance across datasets for TARa@5 and TARr@5 metrics. TARa@5 demonstrates much higher medians overall as expected. Fine-tuned GPT-3.5 Turbo has non-100\% values, but the median remains 100\%.

\begin{figure*}[t]
    \centering    \includegraphics[width=\textwidth]{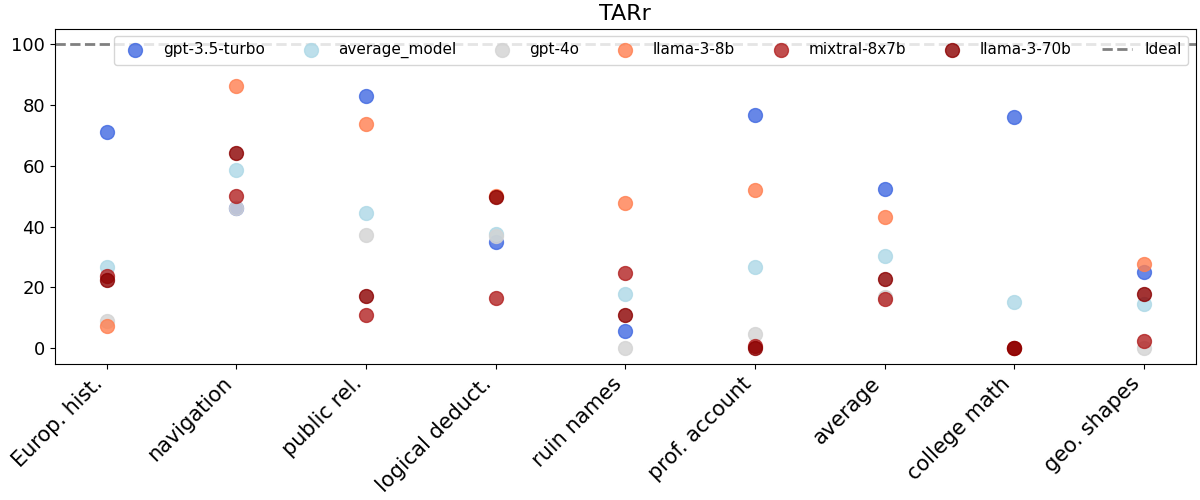}
    \caption{TARr@10 for each model %, in terms of \%. All models are prompted 
    in the few-shot setting. %When the colors change from dark red to dark blue, TARr@5 gets better.
    Dataset colors have been chosen to distinguish them by relatively challenging (increasingly dark red hues) versus relatively easy (increasingly dark blue hues).
    }
    \label{fig:TARr_medians}
\end{figure*}

Figure \ref{fig:TARr_medians} shows the TARr@10 for each task and model in a few-shot setting (for zero-shot scores, see Figure \ref{fig:TARr_medians_0shot} in Appendix \ref{appendix:zero_shot}). GPT-3.5 Turbo %outperforms other models across tasks.
has lower TARr (less instability) than other models, while Llama-3-70B often has very low TARr.

\begin{figure*}[t]
    \centering
    \includegraphics[width=\textwidth]{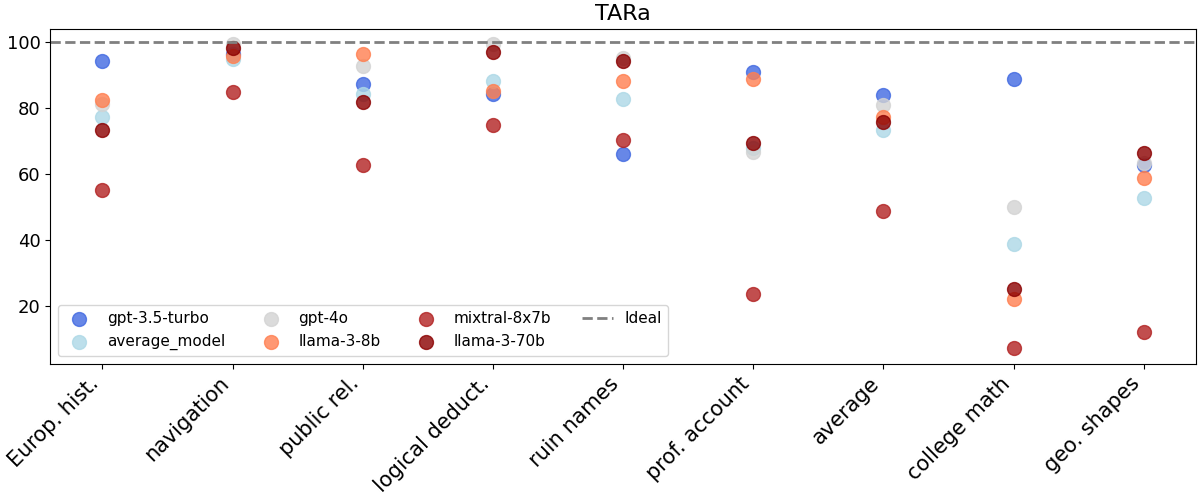}
    \caption{TARa@10 for each task %, in terms of \%. All models are prompted 
    in the few-shot setting. %When the colors change from dark red to dark blue, TARa@5 gets better.
    Models colors have been chosen to distinguish them by relatively low performing (increasingly dark red hues) versus relatively high performing (increasingly dark blue hues).
    }
    \label{fig:TARa_medians}
\end{figure*}

\begin{figure*}[t]
    \centering    \includegraphics[width=\textwidth]{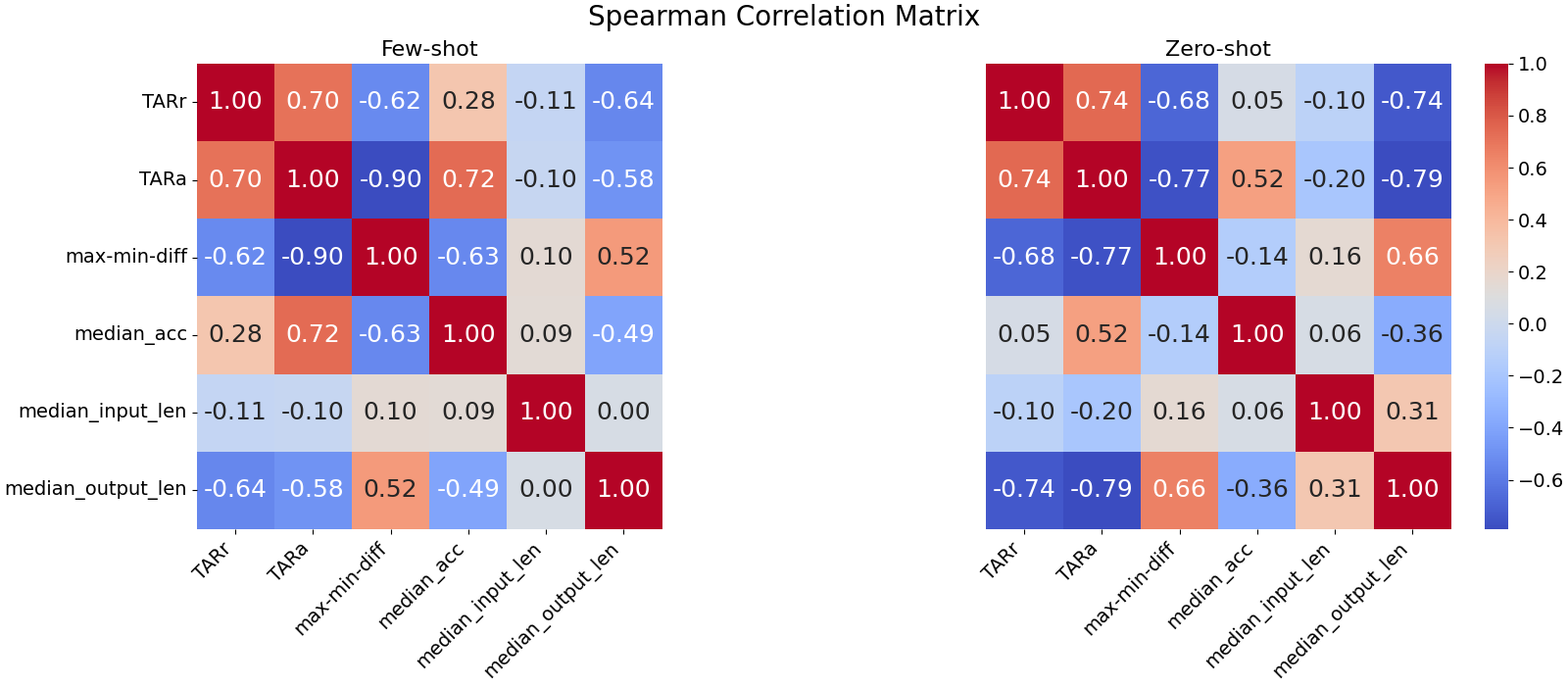}
    \caption{Spearman correlation matrix between metrics in few-shot setting (on the left) and zero-shot setting (on the right).}
    \label{fig:corr_analysis_fewshot}
\end{figure*}

% \begin{figure}[t]
%     \centering    \includegraphics[width=0.7\textwidth]{corr_heat_map0shot.png}
%     \caption{Spearman correlation matrix between metrics in zero-shot setting for the models that show significant instability (all models except the fine-tuned model)}
%     \label{fig:corr_analysis_0shot}
% \end{figure}

Figure \ref{fig:TARa_medians} shows TARa@10 for each condition in a few-shot setting (see Figure \ref{fig:TARa_medians_0shot} in Appendix \ref{appendix:zero_shot} for zero-shot). %Extracting the answer and comparing those result in higher numbers, but 
While the TARa@10 results show less instability than TARr@10 they are still far from perfect and show very task-specific behavior. The high-performance for the navigation task indicate that leaderboards on this task can be expected to be more reliable. On the other hand, the more scattered results for the college math and professional accounting tasks indicate that results reported on these tasks might not be as robust.

% \begin{figure*}[t]
%     \centering
%     \includegraphics[width=0.7\textwidth]{corr_heat_mapfewshot.png}
%     \caption{Spearman correlation matrix between metrics in few-shot setting for the models that show significant instability (all models except the fine-tuned model)}
%     \label{fig:corr_analysis_fewshot}
% \end{figure*}

\subsection{Correlation Analyses}

We perform a Spearman rank correlation analysis on all pairs of the following metrics: TARa@10, TARr@10, maximum minimum accuracy difference, median accuracy, median input length, and output length. Heat map results are shown in %a heat map in 
Figure \ref{fig:corr_analysis_fewshot} for the few-shot and zero-shot prompted models. Here we define accuracy as the median accuracy over the 10 runs with the same model and dataset setup. Input length and output length are median word counts split by space, calculated over the input and output of each LLM experiment setup. We split the words by space instead of using a particular tokenizer, as the models we experimented with use different tokenizers. 

The results show a strong/moderate negative correlation between the output length and TARa@10, as well as between the output length and TARr@10 in few-shot/zero-shot settings. Note this is also consistent with the positive correlation of output length with max-min-diff. %This
These correlations mean that as an LLM's output length increases, the instability of the output increases, resulting in more diverse natural language responses as well as the actual multiple choice answer prediction. %selection. 
The strong negative correlation between LLM output length and instability could motivate those using LLMs commercially to restrict the max generation tokens to control the instability. We also see a strong positive correlation between median accuracy and TARa@10 in the few-shot setting. This indicates that when the LLM is more accurate it becomes more deterministic for multiple choice selections. Additionally, in the few-shot setting, there is a moderate negative correlation between the output length and median accuracy, which indicates that restricting max generation tokens improves both determinisim and accuracy. This is in parallel with the findings in \citep{zhang2024verbosityneqveracitydemystify}.

In addition to general correlations, we also look at correlation maps per model to see how general findings apply to each. We find that all models are more stable when they generate shorter responses. Notably, Mixtral and Llama-3 models are more stable when they are more accurate in few-shot setting but the effect varies in zero-shot setting. Last but not least, GPT-3.5 is more stable when the input is longer in the few-shot setting and the effect is less significant in zero-shot setting. These correlation map figures can be found in Appendix \ref{appendix:corr_matrix}.

\section{Discussion}

Theoretically, at zero temperature the LLM should be deterministic given the same input, with % which results in 100\% TARa@10, 100\% TARr@10, the same BestAcc and WorstAcc, 
values of 100\% for TARa@10 and TAR@10, the same values for BestAcc and WorstAcc,
and 0\% difference in the minimum and maximum values across all tasks. % with hyper-parameters set as they are. 

The TARr@10 scores show that models are not stable at the string variation level but far more stable at the parsed answer level. String variation does not affect a human reader much because we can extract the same answer even if the output format is different, but a downstream system that needs to parse the LLM response can be affected significantly when the format or pattern is different. This should be taken into account when %traditional artificial intelligence (AI) systems are being replaced with LLMs.
using LLMs.

%Other than GPT-3.5 Turbo, the TARr@5 scores put a great deal of pressure on downstream components to be robust to string variation even if the answer is the same in the LLM response payload. This variation doesn't impact humans as much since we have very robust parsing and answer recovery skills, but a downstream system that has to parse the LLM payload would likely never be completely confident that the answer would be recovered--this use-case is increasing in popularity as LLMs replace traditional AI technologies due to their increased performance and ease of use. 

TARa@10 values are %significantly 
much more consistent than TARr@10, % which is the basis of our reported accuracy variation, but 
yet still lead to high instablity of up to 15\% shown in Figure~\ref{fig:acc_diff}. One caveat is that
our answer extraction system has many hard-coded parts, which reduces the generality of the system. Therefore, %and 
we have no guarantee that raw outputs %will retain previous patterns.
will lead to the exactly the same results for our various accuracy metrics, if the experiments are repeated.
%The TARa@5 scores still rarely achieve 100\% but all remain above 88\% so there is better news with the introduction of normalization via answer extraction. However, our answer parsing was hand-developed to cover all the responses seen so far with no guarantee that further raw payloads might diverge outside the scope of the already handled cases.

The maximum-minimum accuracy difference should be 0\% theoretically.  % with that property. 75.0, 31.5, 03.0
All models demonstrate considerable variation on this metric. Mixtral-8x7b on college math is 72\% (75\% - 3\%) for a particularly bad example on suggesting a truly random element in the generative process driving the minimum value to 0\%. This instability lowers confidence in % questions 
the reliability of reporting only a single number in LLM benchmarks. We encourage reporting maximum-minimum scores across runs to have a more robust comparison of models. 

% \begin{table}[t]
%  \begin{center}
%    \begin{adjustbox}{width=0.5\textwidth}
% \begin{tabular}{|l|c|c|}
% \hline \textbf{Model} & \textbf{TARa@5 Median} & \textbf{TARr@5 Median}  \\ \hline
% fine-tuned-gpt3.5-few & 100\% & 100\% \\
% gpt3.5-few & 99\% & 92\% \\
% gpt3.5-0shot & 99\% & 99\% \\
% mixtral8-7b-few & 89\% & 61\% \\
% mixtral8-7b-0shot & 99\% & 88\% \\
% llama8b-few & 92\% & 37\% \\
% llama8b-0shot & 100\% & 98\% \\
% llama70b-few & 93\% & 20\% \\
% llama70b-0shot & 69\% & 2\% \\
% gpt4o-few & 94\% & 3\% \\
% gpt4o-0shot & 93\% & 11\% \\
% \hline
% \end{tabular}
% \end{adjustbox}
% \end{center}
% \caption{\label{tab:model_rankings} Median of TARa@5 and TARr@5 over datasets. The models with the ``few'' in their name are prompted in few-shot setting.}
% \end{table}

\subsection{Implications for Practical Engineering}

%Until the increasing use of multiple GPUs which introduced one source of non-determinism 

Although the use of multiple GPUs introduces some randomness \citep{NondeterminisimGPU}, it can be eliminated by setting random seeds, so that AI models are deterministic given the same input. In that case, %any mistakes 
performance errors could be attributed to the model's generalization capability (e.g., under-/over-fitting). However, engineering optimizations to run LLMs faster, such as continuous batching, chunk prefilling, or prefix caching, might lead to non-deterministic behavior. Since many of the the models are close-sourced (GPT-3-5, GPT-4o), and all are hosted behind APIs we don't control, we can only speculate about the reason for this behavior. In order to support this line of reasoning, we ran Llama3-8b on our local GPUs without any optimizations, yielding % and it led to 
deterministic results. This indicates that the models and GPUs themselves are not the only source of non-determinism.
% While AI models made mistakes and had varied performance over time, it was a function of novel combinations of input data that exposed limitations of the AI model's under-/over-fitting, and failure to represent the data-generating process or other external elements. 

Additionally, we fine-tuned gpt-3.5 using two-fold cross validation. Although, the results indicate that fine-tuning helps %improve determinism 
reduce instability, we %are not completely sure 
hypothesize that a fine tuned model cannot be shared across users and as such our tasks were the only ones being run.  % The actual reason 
%it might be that when a fine-tuned model is hosted on API providers' servers, there is less interaction between a user's API requests and those of other users. %API requests. Hence, we leave the effect of fine-tuning on determinism as future work.

Non-deterministic AI brings new challenges to developers, especially in commercial applications:

\begin{itemize}
    \item The usage of unit tests for AI functions %in the same way as a mathematical function 
    is limited because of non-determinism. %Those unit tests might fail now and have to be handled in a non-unit test framework such as regression testing that tolerates variability. 
    An alternative might be regression testing that tolerates variability.
    \item Low stability might also increase the potential for inexplicable errors that are very different from human mistakes such as responding as ``none of the above'' when the task is a multiple choice selection.
    \item Instability of the format of the outputs can result in downstream parser failures.
    \item Assuming that errors due to non-determinism and errors due to system performance are IID (independently, identically distributed), %the factors are 
    together they produce multiplicative performance reductions. For instance, the performance of a dialog system that uses non-deterministic LLM classifiers to manage transitions would degrade with each additional state in the sequence of dialog states. Specifically, %e.g., 
    a dialog system with 4 classifiers that are 95\% stable will show \(.95^{4} = .814 \) expected performance before even factoring in the accuracy of the classifier on novel inputs.
    \item One of the most important effects is in system complexity that has to handle gracefully ``usually correct but this time wrong'' results. Zipfian distributions are commonly seen in applied AI systems where the frequency of an input/category is inversely related to its rank in count sorted order, \(frequency \propto 1/rank\). Testing tends to concentrate on the frequent events and that delivers confidence that the resulting system is stable for the common inputs. However, the lack of stability undermines the entire foundation of this confidence, especially if mistakes are costly.
\end{itemize}

\section{Conclusion}
We have made a systematic analysis of the determinism of LLMs with the hyper-parameters that should maximize it. Our results show that LLMs can be very non-deterministic in standard setups. Furthermore, an LLM rarely produces the same response ten times given the same input; however, the parsed answer is often more stable. %Another interesting finding is that the accuracy values that you get over different runs are not normally distributed. 
The observation that instability results are not normally distributed makes it more difficult to measure the resulting uncertainty.
Lastly, %the model's capability on a particular task and output length of the model are important factors affecting the stability.
instability is highly variable across tasks for the same model, and across models for the same task.

%\section{Future Work}

%There are various potential directions to build on these findings. 
Other questions about instability remain to be explored.
For instance, how can we %improve 
reduce the instability of LLMs during training or inference time (e.g., adding a meta prompt to indicate the model is only allowed to answer with a single letter)? Second, how %to take the lack of stability 
can the instability of LLMs be taken into account in business products? Third, how should we communicate with decision-makers about instability? Last but not least, more analysis could be done to see if there is any correlation between the stability and %fine-grained errors of the model (e.g., false negatives, false positives) besides correct predictions. 
specific types of errors, such as false positives and false negatives.

\bibliography{colm2025_conference}
\bibliographystyle{colm2025_conference}

\appendix
\section{Appendix}
\subsection{Correlation Matrices Per Model} \label{appendix:corr_matrix}

\begin{figure*}[t]
    \centering    \includegraphics[width=\textwidth]{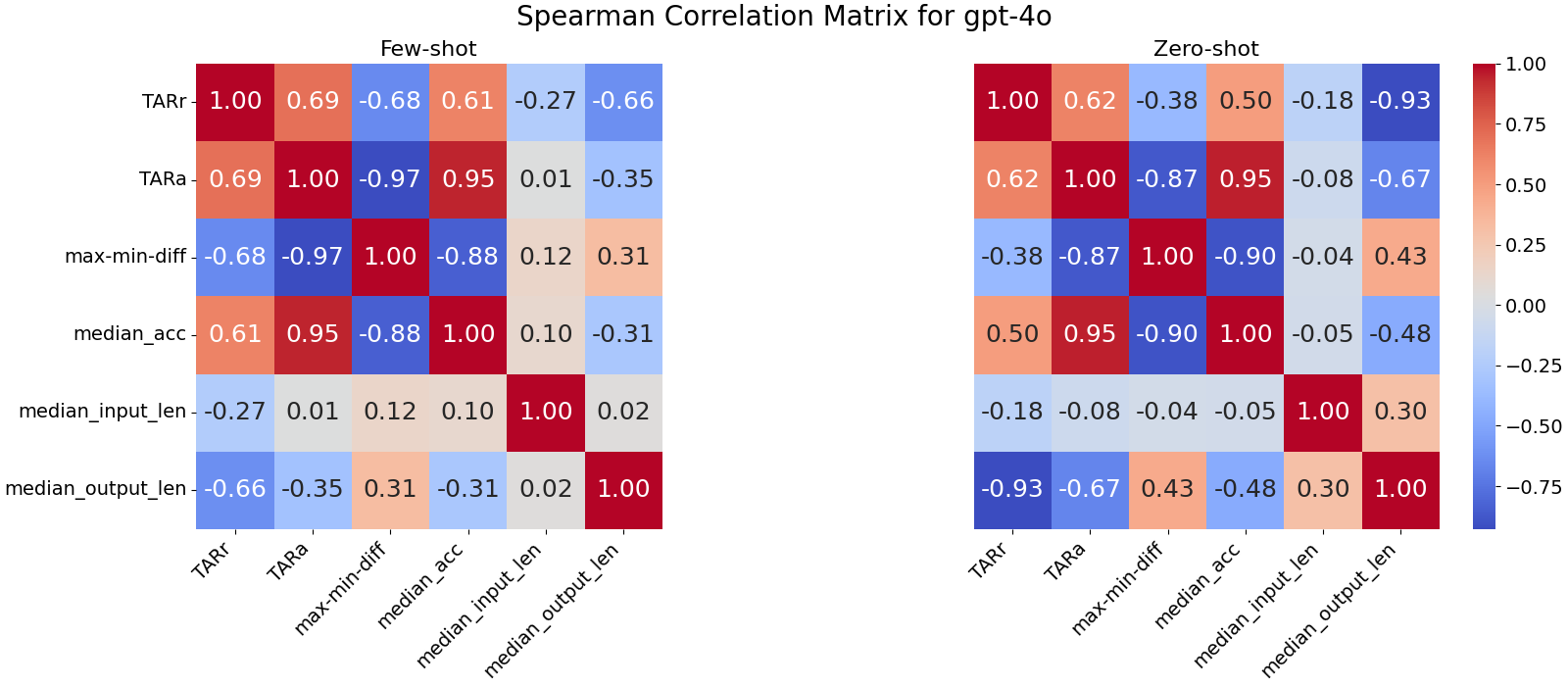}
    \caption{Spearman correlation matrix for GPT-4o between metrics in few-shot setting (on the left) and zero-shot setting (on the right).}
    \label{fig:corr_analysis_fewshot_4o}
\end{figure*}

\begin{figure*}[t]
    \centering    \includegraphics[width=\textwidth]{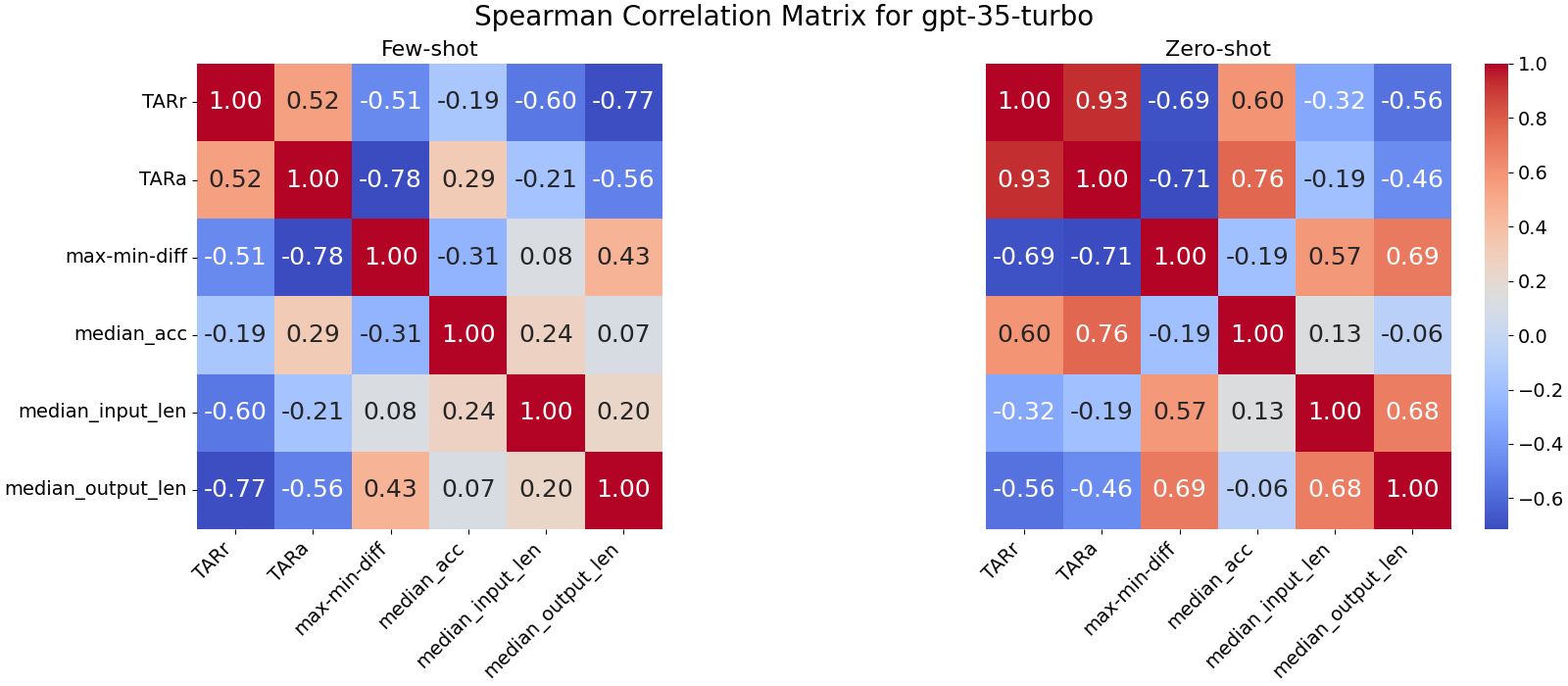}
    \caption{Spearman correlation matrix for GPT-3.5-turbo between metrics in few-shot setting (on the left) and zero-shot setting (on the right).}
    \label{fig:corr_analysis_fewshot_35}
\end{figure*}

\begin{figure*}[t]
    \centering    \includegraphics[width=\textwidth]{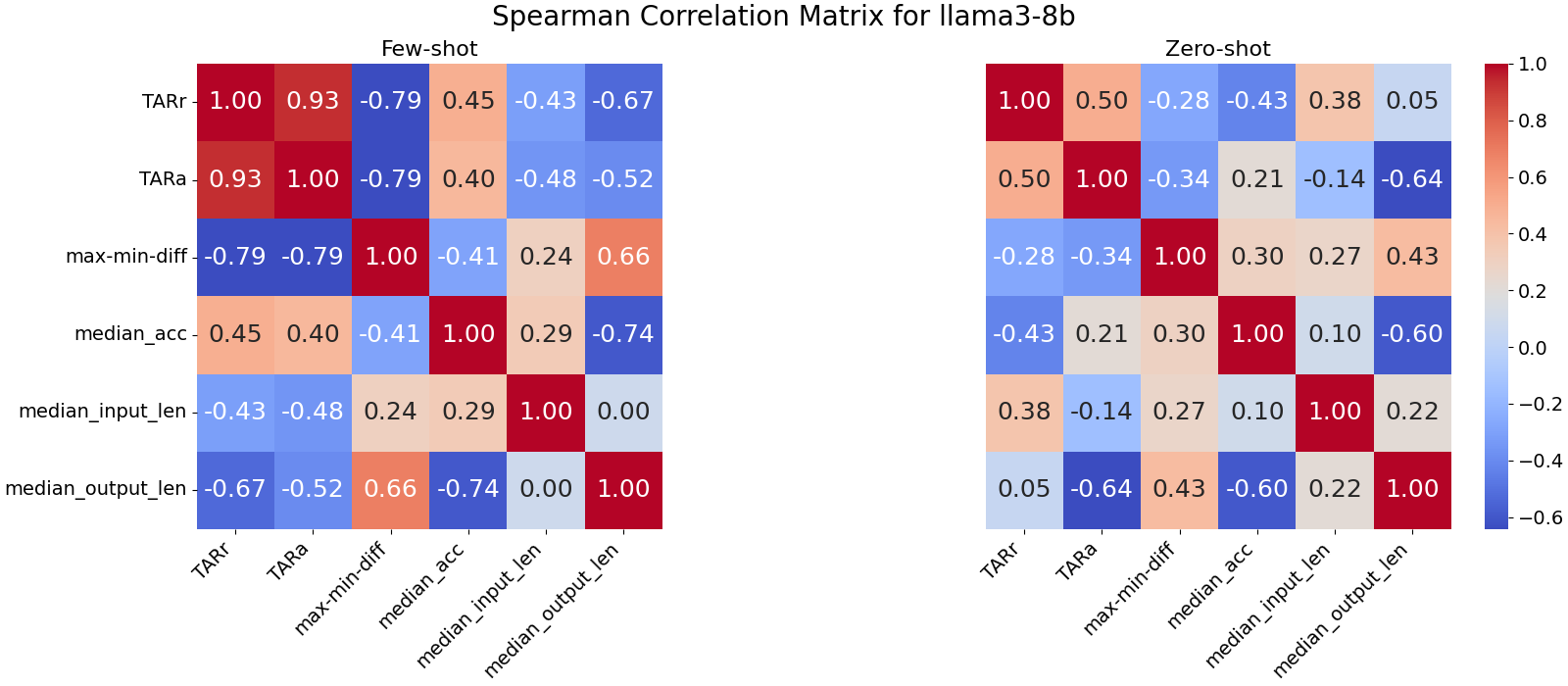}
    \caption{Spearman correlation matrix for Llama-8b between metrics in few-shot setting (on the left) and zero-shot setting (on the right).}
    \label{fig:corr_analysis_fewshot_8b}
\end{figure*}

\begin{figure*}[t]
    \centering    \includegraphics[width=\textwidth]{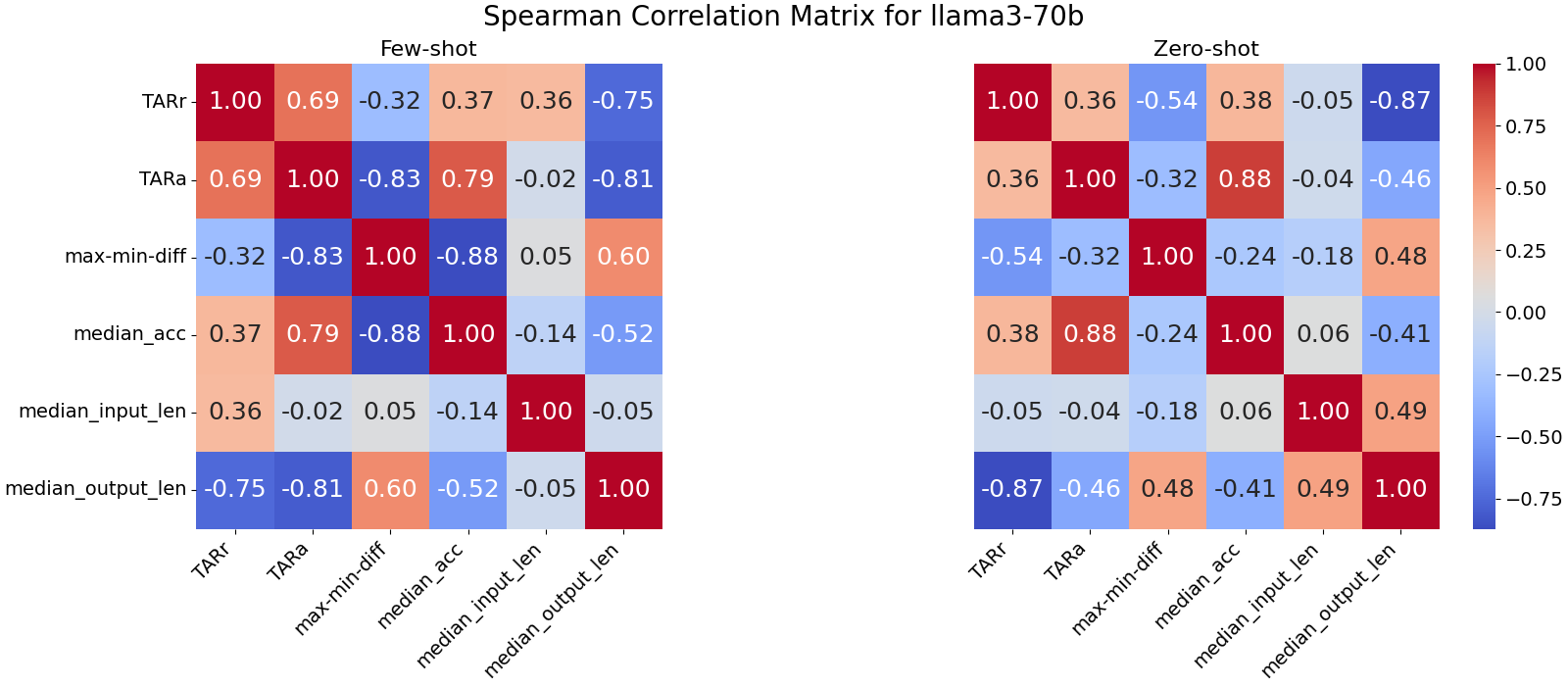}
    \caption{Spearman correlation matrix for Llama-70b between metrics in few-shot setting (on the left) and zero-shot setting (on the right).}
    \label{fig:corr_analysis_fewshot_70b}
\end{figure*}

\begin{figure*}[t]
    \centering    \includegraphics[width=\textwidth]{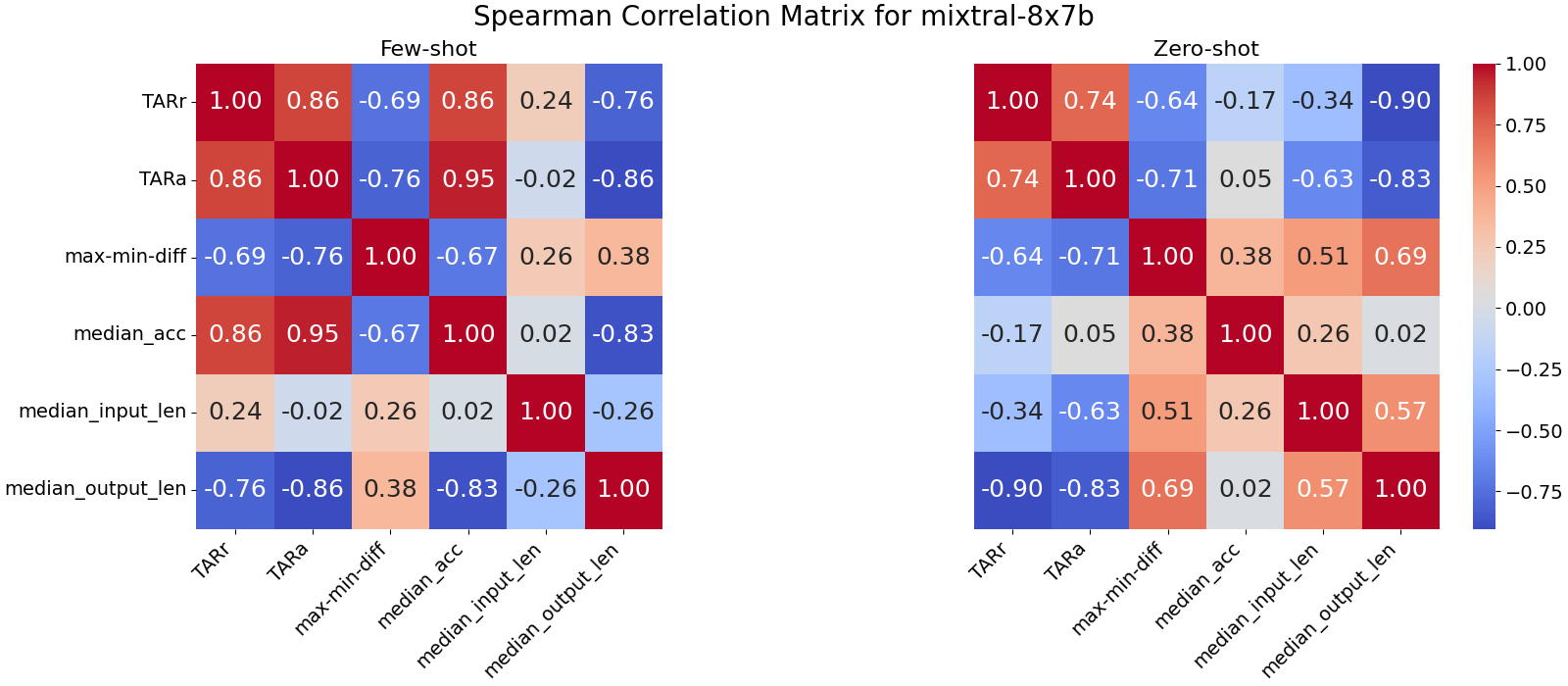}
    \caption{Spearman correlation matrix for Mixtral-8x7b between metrics in few-shot setting (on the left) and zero-shot setting (on the right).}
    \label{fig:corr_analysis_fewshot_mixt}
\end{figure*}

\subsection{Zero-shot Results} \label{appendix:zero_shot}
\begin{table*}[t]
 \begin{center}
   \begin{adjustbox}{width=\textwidth}

 \begin{tabular}{l|c|c|c|c|c}
\hline \textbf{Task} & \textbf{gpt3.5} & \textbf{gpt4o}  & \textbf{llama8b} & \textbf{llama70b} & \textbf{mixtral8-7b}  \\ \hline 
\multicolumn{6}{c}{\textbf{Accuracy Results}} \\\hline
navigation & 67.2, 64.8, 61.6 & 94.8, 92.0, 88.8 & 88.4, 73.0, 54.0 & 94.0, 88.0, 78.4 & 66.0, 57.6, 48.0  \\
geo. shapes & 16.8, 15.2, 13.6 & 76.0, 56.8, 30.4 & 24.4, 18.8, 12.0 & 44.4, 21.6, 6.4 & 29.6, 27.0, 24.8  \\
logical deduct. & 52.8, 50.8, 48.8 & 100.0, 98.6, 96.0 & 72.4, 62.8, 55.6 & 95.6, 92.2, 87.6 & 70.0, 59.6, 49.6  \\
public rel. & 66.4, 65.0, 61.8 & 81.8, 75.5, 66.4 & 28.2, 25.0, 19.1 & 39.1, 26.4, 13.6 & 57.3, 46.8, 35.5  \\
Europ. hist. & 75.2, 74.5, 72.7 & 76.4, 65.2, 55.2 & 38.8, 34.2, 30.3 & 41.2, 27.9, 19.4 & 66.1, 56.1, 45.5  \\
ruin names & 67.2, 65.6, 65.2 & 85.2, 83.2, 80.0 & 54.8, 50.6, 45.6 & 67.6, 60.0, 51.2 & 38.0, 34.4, 30.4  \\
prof. account & 60.3, 53.2, 47.5 & 84.0, 72.0, 58.5 & 36.2, 29.1, 25.5 & 54.6, 38.7, 24.8 & 42.9, 28.9, 20.2  \\
college math & 54.0, 32.0, 15.0 & 85.0, 59.0, 41.0 & 55.0, 34.0, 17.0 & 77.0, 58.0, 40.0 & 57.0, 31.5, 13.0  \\

\hline  
\multicolumn{6}{c}{\textbf{TAR Results}} \\\hline
navigation & 94.4, 94.4 & 91.6, 15.2 & 65.2, 9.2 & 83.2, 4.8 & 77.6, 3.2  \\
geo. shapes & 91.6, 91.6 & 45.6, 0.8 & 60.4, 31.2 & 39.2, 5.6 & 90.4, 83.6  \\
logical deduct. & 92.8, 90.4 & 96.8, 7.6 & 80.4, 37.6 & 92.0, 16.4 & 74.4, 14.0 \\
public rel. & 92.7, 86.4 & 83.6, 38.2 & 82.7, 46.4 & 56.4, 0.9 & 61.8, 10.0 \\
Europ. hist. & 94.5, 94.5 & 74.5, 17.0 & 77.6, 41.2 & 53.9, 6.1 & 63.6, 19.4 \\
ruin names & 95.6, 97.2 & 93.6, 27.6 & 86.8, 26.8 & 79.2, 11.6 & 82.4, 20.8 \\
prof. account & 81.9, 49.3 & 71.3, 4.3 & 77.0, 44.0 & 57.8, 2.1 & 48.2, 4.3  \\
college math & 46.0, 10.0 & 50.0, 0.0 & 45.0, 3.0 & 54.0, 0.0 & 29.0, 2.0 \\
\hline
\end{tabular}
\end{adjustbox}
\end{center}
\caption{\label{tab:general_results_0shot} BestAcc, Median Accuracy, WorstAcc on top; TARa@10, TARr@10 on bottom, for the zero-shot conditions. Results are in terms of percentages.} %These models are prompted in few-shot setting.}
\end{table*}

\begin{figure*}[t]
    \centering
    \includegraphics[width=\textwidth]{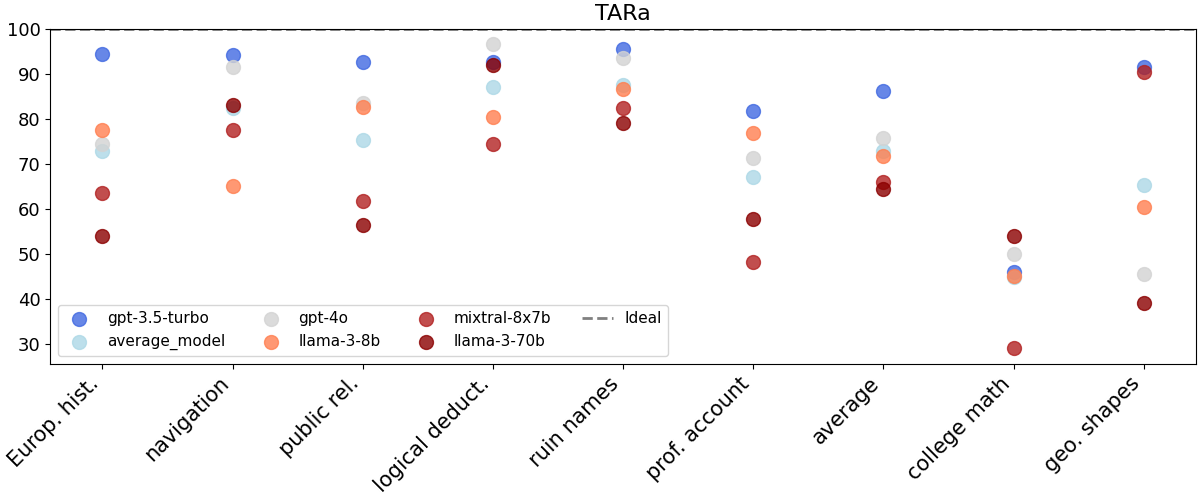}
    \caption{TARa@10 for each task %, in terms of \%. All models are prompted 
    in the zero-shot setting. %When the colors change from dark red to dark blue, TARa@5 gets better.
    Models colors have been chosen to distinguish them by relatively low performing (increasingly dark red hues) versus relatively high performing (increasingly dark blue hues).
    }
    \label{fig:TARa_medians_0shot}
\end{figure*}

\begin{figure*}[t]
    \centering    \includegraphics[width=\textwidth]{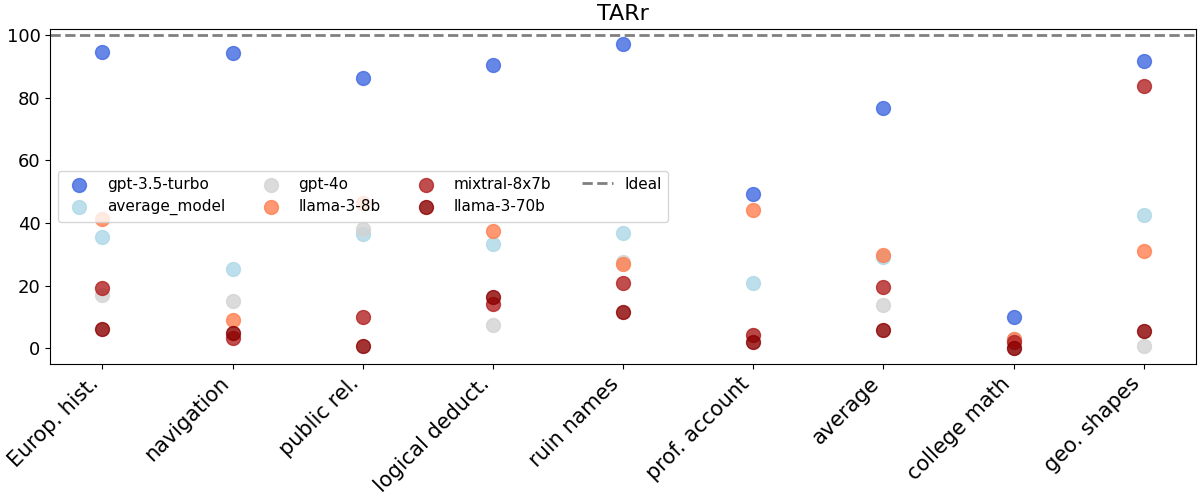}
    \caption{TARr@10 for each model %, in terms of \%. All models are prompted 
    in the zero-shot setting. %When the colors change from dark red to dark blue, TARr@5 gets better.
    Dataset colors have been chosen to distinguish them by relatively challenging (increasingly dark red hues) versus relatively easy (increasingly dark blue hues).
    }
    \label{fig:TARr_medians_0shot}
\end{figure*}

\end{document}